\documentclass[10pt]{article} 
\usepackage[preprint]{rlc}

\usepackage{amssymb}            
\usepackage{mathtools}          
\usepackage{mathrsfs}           
\mathtoolsset{showonlyrefs}     
\usepackage{graphicx}           
\usepackage{subcaption}         
\usepackage[space]{grffile}     
\usepackage{url}                

\usepackage{siunitx}
\usepackage{amsmath}
\usepackage{float}
\usepackage{multicol}
\DeclareMathOperator*{\argmax}{arg\,max}

\title{Bounding-Box Inference for Error-Aware Model-Based Reinforcement Learning}

\author{Erin J. Talvitie \\
    erin@cs.hmc.edu \\
    Harvey Mudd College 
    \And
    Zilei Shao \\
    zoshao@hmc.edu \\
    Harvey Mudd College
    \And
    Huiying Li \\
    erli@hmc.edu \\
    Harvey Mudd College
    \And
    Jinghan Hu \\
    ahu@hmc.edu \\
    Harvey Mudd College
    \And
    Jacob Boerma \thanks{Currently at Tufts University} \\
    jboerma@hmc.edu \\
    Harvey Mudd College
    \And
    Rory Zhao \\
    szhao@hmc.edu \\
    Harvey Mudd College
    \And
    Xintong Wang \\
    xinwang@hmc.edu \\
    Harvey Mudd College}

\begin{document}

\maketitle

\begin{abstract}
 In model-based reinforcement learning, simulated experiences from the learned model are often treated as equivalent to experience from the real environment. However, when the model is inaccurate, it can catastrophically interfere with policy learning. Alternatively, the agent might learn about the model's accuracy and selectively use it only when it can provide reliable predictions. We empirically explore model uncertainty measures for selective planning and show that best results require distribution insensitive inference to estimate the uncertainty over model-based updates. To that end, we propose and evaluate {\em bounding-box inference}, which operates on bounding-boxes around sets of possible states and other quantities. We find that bounding-box inference can reliably support effective selective planning.
\end{abstract}

\section{Introduction}
\label{sec:intro}

A model-based reinforcement learning (MBRL) agent learns a predictive model of its environment, and uses it to inform its decision-making process. Recent successful applications of MBRL approaches such as MuZero \citep{schrittwieser2020mastering} and Dreamer \citep{okada2021dreaming,okada2022dreamingv2,wu2023daydreamer} illustrate the promise of model-based learning as a path toward more capable and more sample-efficient agents. That said, the promise is not yet fully realized. MBRL approaches still tend to be brittle and highly sensitive to model errors.

Acknowledging this, some recent approaches have modified the objectives of model learning to better align with the needs of planning, for instance by focusing on multi-step accuracy \citep[e.g.][]{oh2015action,talvitie2017self} or on accurate rewards and/or state-action values \citep[e.g.][]{grimm2020value}. Ultimately, however, these approaches still rely heavily on the model's accuracy, even if the definition of ``accurate'' is somewhat altered. We must confront the fact that a practical, resource-limited agent cannot always be relied upon to make sufficiently accurate predictions to support planning.

We focus on {\em selective planning}, where the agent estimates the model's input-conditional accuracy and selectively uses the model when it is accurate. In this paper we empirically explore potential uncertainty measures for selective planning and, based on the findings, introduce {\em bounding-box inference}, a novel method for measuring uncertainty over model-based updates to the value function.

\section{Problem Setting and Background}
\label{sec:background}

In this section we formalize the problem-setting and algorithms that we explore in our experiments. We consider {\em Markov decision processes} (MDP). The environment's initial state $s_0$ is drawn from a distribution $\mu$. At each step $t$, the environment is in a state $s_t$. The agent selects an action $a_t$ which causes the environment to transition to a new state $s_{t+1}$ sampled from the distribution given by the transition function: $p(s', s, a) = \text{Pr}(S_{t+1} = s' \mid S_t = s, A_t = a)$. The environment also emits a reward $r_{t+1}$ given by the reward function, $r(s_t, a_t)$. Both $p$ and $r$ are unknown to the agent.

A {\em policy} $\pi$ specifies a way to behave in the MDP. Let $\pi(a \mid s)$ be the probability that $\pi$ chooses action $a$ in state $s$. Given a policy $\pi$, the {\em state-action value} of an action $a$ at state $s$, $q_\pi(s, a)$ is the expected discounted sum of rewards obtained by taking action $a$ in state $s$ and executing $\pi$ forever after: $q_\pi(s_t, a_t) = \text{E}\left[\sum_{i=1}^\infty \gamma^{i-1} R_{t+i} \mid S_t = s_t, A_t = a_t\right]$, where $0 \le \gamma \le 1$ is the discount factor and the expectation is over randomness from both the transition and policy distributions. The agent's goal is to find a policy that maximizes the state value $v_\pi(s) = \text{E}[q_\pi(s, A)]$ in all states $s$, though we assume that the agent's limitations may prevent it from learning such an optimal policy.  

In MBRL we typically seek to learn a model $(\hat{p}, \hat{r})$ that approximates the environment. Alternatively, it is also common to learn a {\em deterministic model}, most commonly an {\em expectation model} that, for a given state and action, provides an estimate of the expected next state and reward. For a deterministic model, we will slightly overload notation and let $\hat{p}(s_t, a_t) = s_{t+1}$ be the model's predicted next state. In this paper we assume that the agent is unable to learn a perfectly accurate model.

\subsection{Sources of Model Error}

Multiple factors can cause inaccurate predictions, which can subsequently cause planning failure. 

{\em Aleatoric uncertainty} refers to uncertainty over outcomes due to stochasticity in the environment. In the presence of aleatoric uncertainty, even perfectly accurate expectation models may cause planning failure unless the value function is linear in the state features \citep{wan2019planning}. So, for some applications, models that represent probability distributions may be necessary.

{\em Epistemic uncertainty} refers to uncertainty over the model parameters themselves, due to limited training. It is common to account for epistemic uncertainty by taking a Bayesian perspective. For instance PILCO \citep{deisenroth2011pilco} uses a Gaussian process model which supports Bayesian inference about the distribution over future rewards considering both aleatoric and epistemic uncertainty. Deep PILCO \citep{gal2016improving} adapts PILCO to use neural networks, using dropout and Monte Carlo methods for approximate inference. Other methods train model ensembles to account for the variety of reasonable predictions based on the training data \citep[e.g.][]{osband2018randomized}. Epistemic uncertainty is reduced by training on additional data; with a sufficiently large training set a single ``best'' set of model parameters typically emerges.

{\em Model inadequacy} refers to the potential inaccuracy of the ``best'' set of model parameters, which may be a result of structural assumptions encoded in the model (e.g. PILCO imposed conditional independence over predicted state variables, given the current state) or resource limitations (e.g. a neural network may have too few nodes to represent the underlying function). In this case, even with deterministic dynamics, even with sufficient training to eliminate epistemic uncertainty, the best fitting model may generate inaccurate predictions that cause planning failure.

Model inadequacy can only be reduced by increasing the expressiveness of the model, which may not always be a practical option. In this paper we focus on mitigating the impact of model inadequacy by selectively using the model in regions of the state space where it can make accurate predictions.

\subsection{Model-Based Value Expansion}

In model-based value expansion (MVE) \citep{feinberg2018model}, the agent gathers data in the environment using a behavior policy $\pi_b$ (which may be fixed or changing) while estimating the optimal state-action values. At time $t$, the agent observes state $s_t$, selects action $a_t$ using $\pi_b$, and observes the next state $s_{t+1}$ and reward $r_{t+1}$. MVE uses the model to calculate {\em multi-step temporal difference errors} \citep{sutton1988learning}, or {\em TD errors}. We define the {\em greedy policy} $\pi_g$ to be the policy that, at every state $s$, takes the action $\argmax_a \hat{q}(s, a)$ with probability one. Starting from $s_{t+1}$, MVE uses $\pi_g$ and the learned model $(\hat{p}, \hat{r})$ to extend the agent's experience into the future, sampling the sequence:
\begin{align*}
  s_t, a_t, r_{t+1}, s_{t+1}, a_{t+1}, \hat{r}_{t+2}, \hat{s}_{t+2}, a_{t+2}, \hat{r}_{t+3}, \hat{s}_{t+3}, \ldots, \hat{r}_{t+h}, \hat{s}_{t+h},
\end{align*}
where each $\hat{r}$ and $\hat{s}$ is generated by the model and all actions from time $t+1$ on are generated by $\pi_g$. This simulated sequence can then be used to calculate the $h$-step {\em TD target}:
\begin{align*}
  \hat{\rho}_h(s_t, a_t, r_{t+1}, s_{t+1}) = r_{t+1} + \sum_{i=2}^h \gamma^{i-1} \hat{r}_{t+i} + \gamma^h \max_{a} \hat{q}(\hat{s}_{t+h}, a).
\end{align*}
The MVE algorithm calculates TD targets at multiple planning horizons up to some maximum horizon $h$ and moves the current state-action value estimate for $s_t, a_t$ toward the average of the targets, which we can more generally consider a weighted average:
\begin{align*}
  \hat{q}(s_t, a_t) \mathrel{+}= \alpha \left(\frac{1}{\sum_{i=1}^h w_i}\left(w_1\rho_1(s_t, a_t, r_{t+1}, s_{t+1}) +  \sum_{i=2}^h w_i\hat{\rho}_i(s_t, a_t, r_{t+1}, s_{t+1})\right) - \hat{q}(s_t, a_t)\right).
\end{align*}
where $\alpha$ is a stepsize metaparameter. When $h=1$, this update is equivalent to Q-learning.

\subsection{Selective Model-Based Value Expansion}

The MVE algorithm is particularly amenable to selective planning by adjusting the weights according to the reliability of each TD target. For example, the STEVE algorithm \citep{buckman2018sample} learns an ensemble of models and value functions and bases the MVE weights on the variance of the ensemble of TD targets. If the models disagree, the estimated target should not be trusted.

\citet{abbas2020selective} argue that ensemble variance primarily measures epistemic uncertainty and show that, with sufficient training, an ensemble of models may agree on the ``best'' model, which may still be inaccurate. They argue that model inadequacy may be detected using methods for learning stochastic models in the face of aleatoric uncertainty; model error will be interpreted by the training process as noise and manifest as higher variance in the model's predictions. They train a single model with a Gaussian approximation of the transition distribution and base the weights on the predicted variance, showing that this does indeed mitigate the impact of model inadequacy.

In this paper we follow \citet{abbas2020selective} in using the spread of the model's predictions to detect model inadequacy but follow \citet{buckman2018sample} in measuring uncertainty over TD targets rather than states. We will calculate an uncertainty $u_i$ for each TD target $\hat{\rho}_i$, with $u_1 = 0$. Then MVE weights will be determined by a softmin distribution: $w_i = e^{\frac{-u_i}{\tau}}/\sum_j e^{\frac{-u_j}{\tau}}$, where $\tau$ is the {\em temperature}. When $\tau \rightarrow \infty$, the update approaches MVE, which equally weights all targets. If $u_i > 0$ for $i > 1$, then as $\tau \rightarrow 0$ the update approaches Q-learning.

\section{Experiments with Hand-Coded Models}
\label{sec:oracles}

In this section we introduce a simple illustrative problem designed to distill some issues related to planning with an inadequate model and experiment with idealized hand-coded models. In Section \ref{sec:learned} we experiment with learned models and in Section \ref{sec:moreexp} we consider a less contrived problem\footnote{Source code for all experiments can be found at \url{https://github.com/LACE-Lab/bounding-box}}.

\subsection{The Go-Right Problem}

\begin{figure}
  \centering
  \subfloat{\includegraphics[width=0.45\linewidth]{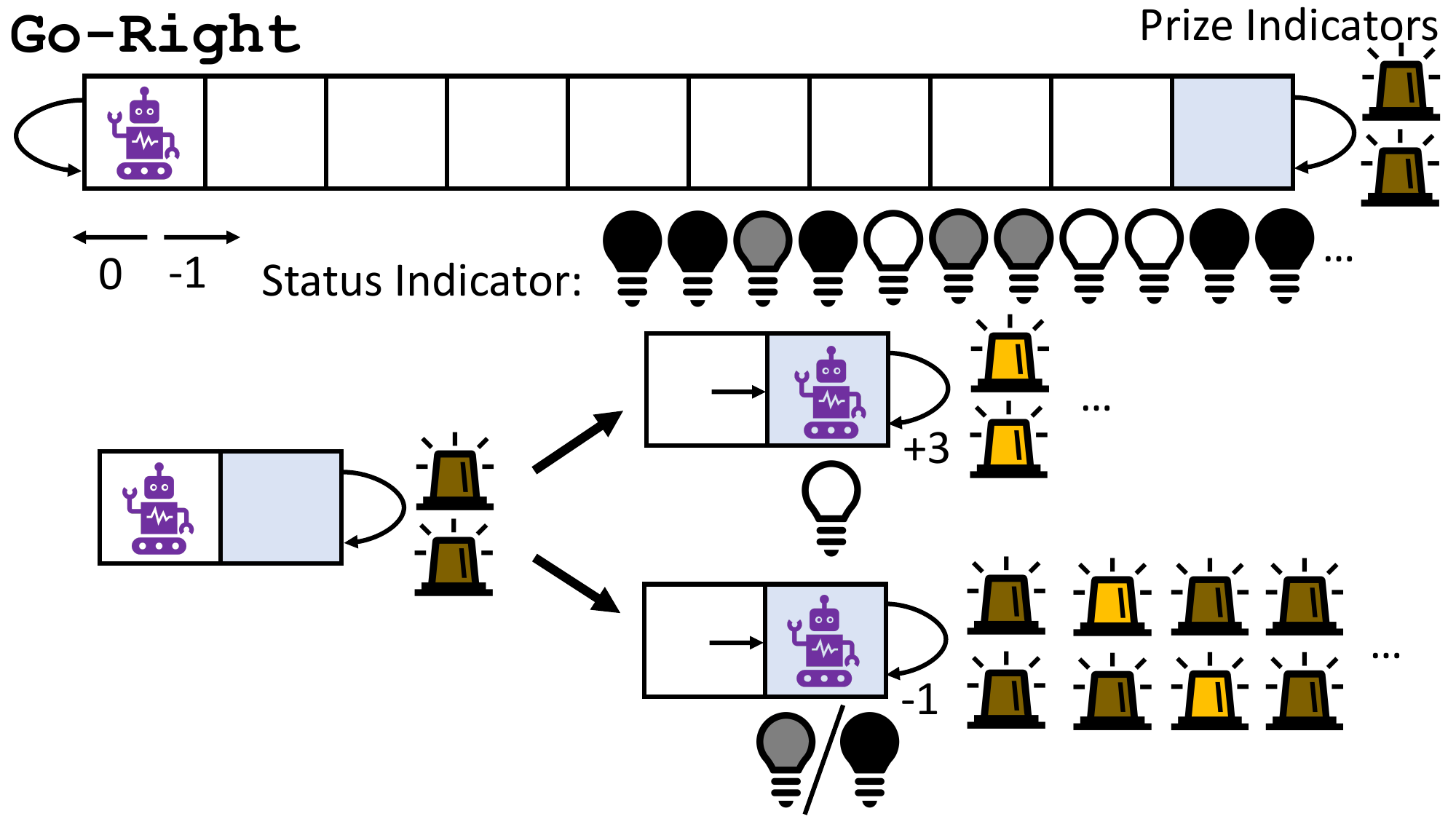}}\qquad
  \subfloat{\includegraphics[width=0.45\linewidth]{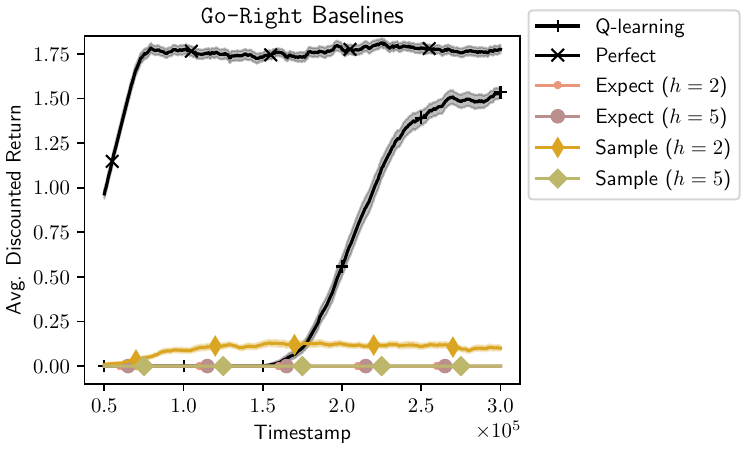}}
  \caption{Left: an illustration of the \texttt{Go-Right} domain. Right: Results of unselective MVE planning in \texttt{Go-Right}. The curves are smoothed so that each point is the average of the previous 100 episode scores. The shaded regions represent the (smoothed) standard error at each point.}
  \label{fig:goright}
\end{figure}

The \texttt{Go-Right} problem is illustrated in Figure \ref{fig:goright} (left). At a high level, the agent begins at the far left of a hallway and must take 10 steps to the other end to receive a prize. Moving right generally gives -1 reward (moving left gives 0 reward). The discount factor $\gamma = 0.9$.

The agent also observes a {\em status indicator} light, which switches between 3 possible intensities in a deterministic but 2nd-order Markov pattern. If the agent enters the prize location at the same time that the status indicator reaches full intensity, the agent wins the prize. Specifically, until the agent leaves the prize location, moving right gives the agent 3 reward, rather than -1. 

The two {\em prize indicator} lights tell the agent whether it has won the prize. They are both off when the agent is not in the prize location and both on when the agent has won the prize. When the agent is in the prize location but has not won the prize, the lights flash in a Markov pattern.

The agent observes continuous variables representing its position and the intensities of the indicator lights. At initialization, each variable is given a random offset, which remains until the problem is reset. As such, though the underlying dynamics are discrete, the data received by the agent is continuous-valued. A more detailed description of the dynamics can be found in Appendix \ref{app:goright}.

\subsection{Experimental Setup}

We approximate an infinite horizon by exposing the agent to \texttt{Go-Right} for 500 steps at a time before truncating the interaction and resetting. The agent does not receive a termination signal, and does not perceive the reset as a transition. We use a uniform-random behavior policy to collect training data. After each 500-step interaction, we evaluate the agent's greedy policy (also for 500 steps).

The $\hat{q}$ function is a lookup table over the (non-Markov) underlying discrete dynamics. Unless otherwise specified, we used $h=5$. For each agent we performed a joint sweep over $\alpha$ and $\tau$ (for details, see Appendix \ref{app:sweep}). We then ran 50 independent trials with the selected metaparameter values.

\subsection{Unselective Planning Results}

We begin by studying MVE planning using idealized hand-coded models that represent reasonable model limitations. The following discussion refers to results shown in Figure \ref{fig:goright} (right), which shows the discounted sum of rewards obtained by the learned greedy policy, averaged over 50 trials.

First note that, despite the partially observable environment, Q-learning is able to learn a good policy that goes right and then repeatedly enters the prize location until the agent wins the prize. The agent labeled ``Perfect'' performs unselective MVE using a perfectly accurate model that accounts for the 2nd-order Markov dynamics. We see that planning can significantly improve learning performance and that access to the full state allows the agent to minimize the number of costly {\em right} actions.

\subsubsection{Expectation Models}
\label{sec:expectation}

The agents labeled ``Expect'' perform unselective MVE using a hand-coded, Markov expectation model that gives the exact least squares estimate of the next state. Being Markov, the model cannot accurately predict the deterministic next state. Specifically, when the agent enters the prize location, the model predicts prize indicator values of $\frac{1}{3}$, which is not a value ever observed in the environment. As it cannot be informed by data, the value of $\hat{q}$ given this impossible state is arbitrary, depending on the biases of the function approximator. Our lookup table considers intensities below 0.5 to be ``off'' so the model effectively never predicts that the agent will receive the prize.

When $h = 5$, this issue causes the agent to never learn to go right. One common strategy for mitigating model error is to shorten planning rollouts \citep[e.g.][]{jiang2015dependence,janner2019trust} and $h = 2$ planning does perform slightly better. However, $h=2$ planning still causes policy learning to fail, despite using only a single simulated step from the model.

\subsubsection{Sampling Models}
\label{sec:sampling}

The agents labeled ``Sample'' use a stochastic Markov model of the environment. Given a state and action, the model provides independent samples of the exact maximum-likelihood distribution over each variable of the next state. Note that \texttt{Go-Right} {\em satisfies} the independence assumption; in a deterministic system all variables are independent. However, the model is inaccurate so the maximum-likelihood distributions are not deterministic. When the agent enters the prize location, the model assigns a probability of $\frac{1}{3}$ to each prize indicator light turning on and therefore only probability $\frac{1}{9}$ to both lights turning on and the agent winning the prize.

Because the sampling model assigns a low probability to the agent receiving the prize, these agents do not learn good policies. As above, we see only slightly better performance when $h=2$.

\subsection{One-Step Predicted Variance}

\begin{figure}
  \centering
  \includegraphics[width=\linewidth]{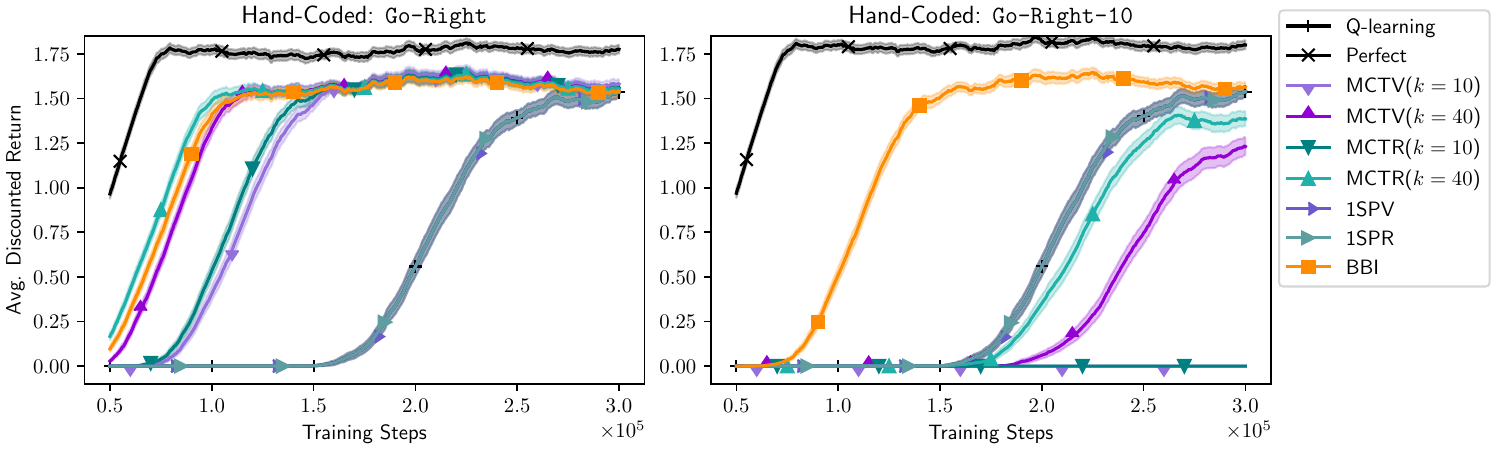}
  \caption{Selective planning with hand-coded models in \texttt{Go-Right} (left) and \texttt{Go-Right-10} (right).}
  \label{fig:oselect}
\end{figure}

We now turn to Figure \ref{fig:oselect} (left), which shows results of selective planning. The agent labeled ``1SPV'' uses the uncertainty measure proposed by \citet{abbas2020selective}, which we call {\em one-step predicted variance (1SPV)}. We use the same hand-coded expectation model as in Section \ref{sec:expectation}, but the model also outputs the exact maximum-likelihood, input-conditional variance $\hat{\sigma}^2_d(s_t, a_t)$ for each dimension $d$ of the next state $s_{t+1}$ as well as the reward $\hat{\sigma}^2_r(s_t, a_t)$. Following \citet{abbas2020selective}, the uncertainty associated with the TD target at horizon $i > 1$ is then $u_i = \sum_{j=0}^{i-1} \left(\sum_d \hat{\sigma}^2_d(s_{t+j}, a_{t+j}) + \hat{\sigma}^2_r(s_{t+j}, a_{t+j})\right)$\footnote{\citet{abbas2020selective} assumed that the reward function was known so did not include the reward term.}. Summing the variances over the state dimensions corresponds to a conditional independence assumption over the variables of the next state. As far as we are aware, summing the variances over rollout steps has no theoretical basis, but it heuristically encodes the intuition that early uncertainty should make later steps uncertain as well.

In \texttt{Go-Right} there is a persistent source of state uncertainty; the status indicator has high predicted variance at every step. However, in most states the status indicator is largely irrelevant to $q$-values, so the 1SPV measure is too conservative. With the best performing temperature, the performance of the learned policy is essentially identical to that of Q-learning. This is preferable to the catastrophic failure of unselective planning in Section \ref{sec:expectation}, but the model is underutilized.

\subsection{Monte-Carlo Target Variance}
\label{sec:mctv}

The primary issues with 1SPV that we have identified are that state uncertainty is not always a good proxy for TD target uncertainty and that the method for computing multi-step uncertainty is not theoretically motivated. We can address both of these via principled inference of the model's uncertainty over {\em TD targets} rather than individual transitions.

As in Deep PILCO \citep{gal2016improving}, the agents marked ``MCTV'' in Figure \ref{fig:oselect} (left) use the Monte Carlo method as a general-purpose inference method to approximate the model's uncertainty over multistep TD targets. They use the same sampling model as in Section \ref{sec:sampling}, but for each planning step they sample $k$ independent rollouts (each beginning with the same initial transition from the environment). This produces $k$ TD targets at each horizon $i$: $\hat{\rho}^1_i,\hat{\rho}^2_i,\ldots,\hat{\rho}^k_i$. The MVE update at horizon $i$ uses the average of the sampled TD targets: $\hat{\rho}_i = \frac{1}{k} \sum_{j=1}^k \hat{\rho}^j_i$. The uncertainty $u_i$ associated with the TD target $\hat{\rho}_i$ is the sample variance of the TD targets: $u_i = \frac{1}{k-1} \sum_{j=1}^k (\hat{\rho}^j_i - \hat{\rho}_i)^2$.

We show results with $k = 10$ and $k = 40$. We observe that selective planning that infers the impact of the model's uncertainty on TD target uncertainty may allow the model to be used effectively, despite persistent state error. The higher sample size predictably performs better. 

\subsubsection{Limitations of Target Variance}

Though intuitive, predicted TD target variance is not inherently the ideal signal for selective planning. Remember that there {\em is no stochasticity in \texttt{Go-Right}}. The model's predicted target variance is a symptom of underfitting and not an estimate of any underlying probabilistic quantity. The variance estimate depends upon the model's structural biases and the observed frequency of transitions, which in turn depends upon the model's representation of the state and the behavior policy.

As an illustration of this issue, we run the same experiment but in a version of \texttt{Go-Right} that has 10 prize indicators instead of 2, which we call \texttt{Go-Right-10}. Though nothing about the underlying task has changed, the sampling model will now assign probability $\frac{1}{3^{10}} = \frac{1}{\num{59049}}$ to the agent receiving the prize when it enters the prize location. Not only does this cause the model to further devalue going right, it also dramatically reduces the model's variance over TD targets that involve this transition.

The results are shown in Figure \ref{fig:oselect} (right). Both MCTV agents now perform worse than Q-learning. We expect that performance would improve with more accurate inference, for instance via more Monte Carlo samples, but note that we can further alter this problem to reduce the model's target variance arbitrarily while the model inadequacy remains the same. As such, even if exact inference were available, the magnitude of the predicted target variance would not be a reliable indicator of model inadequacy. Any non-zero TD target variance could signal catastrophic planning failures.

\subsection{Monte Carlo Target Range}

In order to reduce sensitivity to the model's learned probability distribution we explore an alternative measure of the spread of TD targets: the {\em range}, i.e. the difference between the maximum and minimum possible TD targets\footnote{If the model's distribution over TD targets is not bounded, one could instead consider measuring a quantile slightly below 1 in place of the max and a quantile slightly above 0 in place of the min.}. If the range is small, then all possible model-generated TD targets are similar. If the range is large, then the model is unsure about the TD target.

The agents labeled ``MCTR'' in Figure \ref{fig:oselect} use the same Monte Carlo procedure as in Section \ref{sec:mctv} but instead of calculating the variance of the TD targets, the uncertainty for horizon $i$ is $u_i = \max_j \hat{\rho}^j_i - \min_j \hat{\rho}^j_i$. In \texttt{Go-Right} we see that at both $k = 40$ and $k = 10$ the MCTR agents outperform the MCTV agents. In \texttt{Go-Right-10}, MCTR with $k=40$ outperforms MCTV, suggesting that target range is more robust to the change in probability distribution. That said, it still performs worse than Q-learning because in some rollouts at least one of the extreme values is not sampled. As such, even the MCTR agent is somewhat sensitive to the model's predicted probability distribution.

For completeness, we also include a one-step prediction range (1SPR) agent, which is the same as 1SPV except it sums the range of each state variable rather than the variance. Unsurprisingly, it performs essentially the same as the 1SPV agent.

\subsection{Bounding-Box Inference}

To further reduce sensitivity to the predicted distribution, we introduce a light-weight alternative inference method that infers ranges over TD targets from bounds over one-step predictions. We hand-coded a Markov model that operates on a {\em bounding-box} over states. Let $\underline{s}$ be a vector of minimum values for each dimension, $\overline{s}$ contain maximum values, and $\overline{\underline{s}}$ be the corresponding bounding-box. The model also operates on a set of actions, which, for notational consistency, we represent with $\overline{\underline{a}}$.

The hand-coded model takes an input bounding-box and outputs the exact bounds for each dimension of the next states and reward: $\overline{\underline{p}}(\overline{\underline{s}}_t, \overline{\underline{a}}_t) = \overline{\underline{s}}_{t+1}$ and $\overline{\underline{r}}(\overline{\underline{s}}_t, \overline{\underline{a}}_t) = \overline{\underline{r}}_{t+1}$. We also modify $\hat{q}$ to operate on bounding-boxes: let $\overline{q}(\overline{\underline{s}}_t, \overline{\underline{a}}_t) = \sup_{s \in \overline{\underline{s}}_t, a \in \overline{\underline{a}}_t}\hat{q}(s, a)$ and $\underline{q}(\overline{\underline{s}}_t, \overline{\underline{a}}_t) = \inf_{s \in \overline{\underline{s}}_t, a \in \overline{\underline{a}}_t}\hat{q}(s, a)$

We now describe how to select a set of greedy actions for a bounding-box of states. We infer an upper bound on the value of behaving greedily: $\overline{v}(\overline{\underline{s}}_t) = \max_a \overline{q}(\overline{\underline{s}}_t, a)$, as this is the maximum possible value that can be obtained by taking one of the included actions in one of the included states. Similarly, the lower bound $\underline{v}(\overline{\underline{s}}_t) = \max_a \underline{q}(\overline{\underline{s}}_t, a)$, because in all included states there is an action that obtains at least this much value. For a bounding-box $\overline{\underline{s}}_t$ we let the greedy action set $\overline{\underline{a}}_t = \{a \mid \overline{q}(\underline{\overline{s}}_t, a) \ge \underline{v}(\underline{\overline{s}}_t)\}$, the actions whose value bounds overlap with the greedy value bounds; these are actions that could potentially be selected greedily within the given state bounding-box. 

Given these components, selective MVE at time $t$ can now perform a {\em bounding-box rollout}, starting with $s_{t+1} = \overline{s}_{t+1} = \underline{s}_{t+1}$, which is generated by the environment. This produces the sequence 
\begin{align*}
  s_t, a_t, r_{t+1}, s_{t+1}, a_{t+1}, \overline{\underline{r}}_{t+2}, \overline{\underline{s}}_{t+2}, \overline{\underline{a}}_{t+2}, \overline{\underline{r}}_{t+3}, \overline{\underline{s}}_{t+3}, \ldots, \overline{\underline{r}}_{t+h}, \overline{\underline{s}}_{t+h}.
\end{align*}

We can use these quantities to infer bounds over the TD target at horizon $i$:
\begin{align*}
  \overline{\rho}_i(s_t, a_t, r_{t+1}, s_{t+1}) = r_{t+1} + \sum_{i=2}^i \gamma^{i-1} \overline{r}_{t+i} + \gamma^{t+i} \overline{v}(\overline{\underline{s}}_{t+i}),
\end{align*}
and the symmetric calculation for $\underline{\rho}_i$. We let the uncertainty at horizon $i$ be the range $u_i = \overline{\rho}_i - \underline{\rho}_i$. 

Note that the inferred bounds may be loose. For instance, we apply the relaxing assumption that all state variables and the reward can independently achieve their extreme values. In the real environment, these variables may have relationships that prevent this. That said, the inferred target range does represent a conservative upper bound for the true target range. Overly conservative uncertainty estimates may, at worst, unnecessarily limit model usage. When the model is confidently incorrect, it may interfere with policy learning. Thus, we err on the side of conservative bounds.

The agent labeled ``BBI'' in Figure \ref{fig:oselect} uses the expectation model described in Section \ref{sec:expectation} to generate TD targets and uses a BBI rollout to generate uncertainties. In \texttt{Go-Right} it performs similarly to the MCTR agent, indicating that the inferred target ranges are roughly as informative as the Monte Carlo estimates. In \texttt{Go-Right-10} we see, as expected, that BBI is unaffected by the change in the model's probability distributions. To ensure that the BBI agent is not simply overestimating the value of going right, we also ran the same experiments with a lower discount factor $\gamma = 0.85$, for which the optimal policy is to always go left. The results in Appendix \ref{app:goleft} show that all of the agents correctly learn a policy that never goes right. Overall, we observe that measures of spread that are less sensitive to the model's predicted distribution may be more robust for selective planning.

\section{Bounding-Box Inference in Learned Models}
\label{sec:learned}

In this section we describe simple procedures for performing bounding-box inference with foundational model classes. For simplicity, we limit our discussion to models with a single dimensional output; the ideas here can be straightforwardly extended to models with multi-dimensional output.

We also perform experiments similar to the above, but we learn models along with the policy. In all cases, the model was tasked with predicting the {\em change} in the state, rather than the value of the next state. Furthermore, to avoid any concerns about learning interference, we learn a separate predictive model for each state dimension and the reward. For each experiment we performed a parameter sweep over $\alpha$ and $\tau$ as before. We did not formally sweep the metaparameters of the model-learning algorithms; we do not aim to learn the best possible model, but rather to make the most of the model that is learned. We provide details about model-learning in Appendix \ref{app:metaparams}.

\subsection{Types of Bounding-Box Queries}

We distinguish between two related but distinct bounding-box queries that require slightly different treatment. An {\em output bound} query asks for bounds over the function's output, reflecting only uncertainty arising from the uncertain input. For example, in the BBI procedure we used output bound queries on $\hat{q}$, computing bounds on the estimated values of the possible states and actions.

In contrast, an {\em outcome bound} query asks for bounds over the possible real outcomes from the environment, incorporating the model's uncertainty. In the BBI procedure we used outcome bound queries for the state and reward predictions. The predicted bounding-box reflects both the set of possible inputs and the model's uncertainty over, e.g., the next value of the status indicator.

\subsection{Linear Models}

A linear model approximates the outcome as a linear combination of features of the input. Let features $\phi_1, \phi_2, \ldots, \phi_n$ be functions of the input such that $\phi_i(x) \in \mathbb{R}$. A linear model has a set of weights $\theta_1, \theta_2, \ldots, \theta_n \in \mathbb{R}$ and the model's prediction $\hat{f}(x) = \sum_{i=1}^n \theta_i \phi_i(x)$. 

\subsubsection{Output Bound Queries}

For each feature $\phi_i$, let $\overline{\phi}_i(\overline{\underline{x}}) = \sup_{x \in \overline{\underline{x}}} \phi_i(x)$ and $\underline{\phi}_i(\overline{\underline{x}}) = \inf_{x \in \overline{\underline{x}}} \phi_i(x)$. To compute bounds on the output, we imagine that all features can independently reach their extreme values. Then it is straightforward to upper bound the output of $\hat{f}$: $\overline{f}(\overline{\underline{x}}) = \sum_{i=1}^n \max(\theta_i, 0) \overline{\phi}_i(\overline{\underline{x}}) + \sum_{i=1}^n \min(\theta_i, 0) \underline{\phi}_i(\overline{\underline{x}})$. That is, we calculate the output assuming that features with positive weight have their maximum values and those with negative weight have their minimum values. We can calculate $\underline{f}(\overline{\underline{x}})$ symmetrically.

It may be possible to obtain tighter bounds with some structural knowledge. For instance, if the features $\phi_i, \phi_{i+1}, \ldots, \phi_{j}$ are known to constitute a {\em one-hot encoding}, where one feature takes the value 1 and the others take the value 0, then their maximum contribution to the output is $\max_{l \in \{i, \ldots, j\}} \theta_l$.

\subsubsection{Outcome Bound Queries}

To account for uncertainty over the true outcome, we must maintain additional statistics that measure the spread of observed outcomes. For example, we could maintain the maximum and minimum residuals ever observed, $\overline{z}$ and $\underline{z}$. Then, given an input bound $\overline{\underline{x}}$, we could estimate an upper bound on the outcome as $\overline{f}(\overline{\underline{x}}) + \overline{z}$. Alternatively, we could estimate input-conditional estimates of spread, for instance by binning the input space and maintaining the extreme residuals in each bin.

\subsection{Regression Trees}

A regression tree \citep{breiman1984classification} defines a piece-wise function as a tree. For simplicity, we assume a binary tree with constant leaf models; each leaf node $l$ predicts the average observed outcome. Each internal node $n$ is associated with a binary feature $\phi_n$. The prediction of node $n$ is recursively either the prediction of the left or right child, depending on the value of $\phi_n(x)$, formally $\hat{f}_n(x) = (1 - \phi_n(x)) \hat{f}_{n.left}(x) + \phi_n(x) \hat{f}_{n.right}(x)$. The overall prediction $\hat{f}(x) = \hat{f}_{root}(x)$.

We assume that each leaf model can support output bound and outcome bound queries (certainly true of constant models). We can leverage the recursive structure of the tree to compute either type of bound query for the tree as a whole. Given an input bound $\overline{\underline{x}}$, at node $n$ either $\underline{\phi}_n(\overline{\underline{x}}) = \overline{\phi}_n(\overline{\underline{x}})$, in which case we can recur to the appropriate child as above, or $\underline{\phi}_n(\overline{\underline{x}}) \ne \overline{\phi}_n(\overline{\underline{x}})$, in which case the predicted bounds must account for the predictions of both children. In the latter case, we let $\overline{f}_n(x) = \max(\overline{f}_{n.left}(\overline{\underline{x}}), \overline{f}_{n.right}(\overline{\underline{x}}))$ and symmetrically $\underline{f}_n(x) = \min(\underline{f}_{n.left}(\overline{\underline{x}}), \underline{f}_{n.right}(\overline{\underline{x}}))$. 

\subsubsection{Experiments}

We learned models using the fast incremental regression tree (FIRT) algorithm \citep{ikonomovska2011learning}, which maintains statistics in each leaf and splits the leaf by adding a feature when confident that doing so would reduce the predicted variance of the outcome. Leafs used constant models, predicting the sample mean $\hat{y}$ of observed outcomes. As outcome bounds, leaves stored the maximum and minimum observed outcome. For sampling, leaves stored the sample variance $\hat{\sigma}^2$ and sampled from $\mathcal{N}(\hat{y}, \hat{\sigma}^2)$. As is typical, features were thresholds over input dimensions. In each leaf, the candidate thresholds for splitting were the observed values of each input dimension.

\begin{figure}
  \centering
  \includegraphics[width=\linewidth]{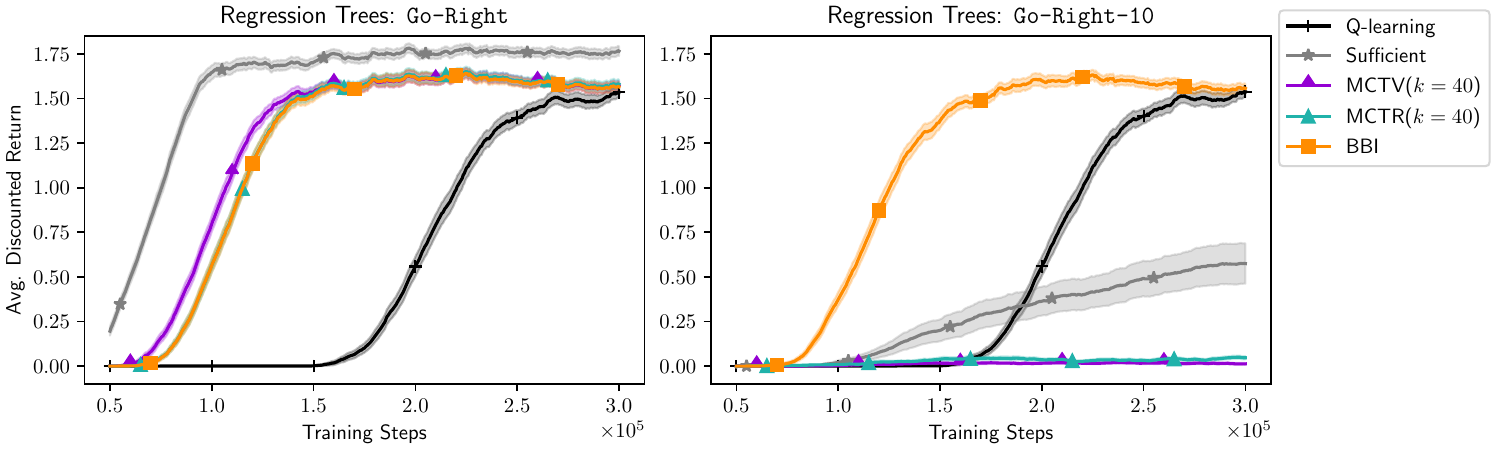}
  \caption{Selective planning with decision tree models in \texttt{Go-Right} (left) and \texttt{Go-Right-10} (right).}
  \label{fig:dtgoup}
\end{figure}

Figure \ref{fig:dtgoup} (left) shows results in \texttt{Go-Right}. The agent marked ``Sufficient'' performs unselective MVE with a learned model that has access to the full 2nd-order Markov state, and is therefore in principle capable of learning a perfectly accurate model. This agent eventually performs similarly to the perfect model agent. All selective planning methods outperform Q-learning, with MCTV performing slightly better than BBI and MCTR. BBI and MCTR perform nearly identically, suggesting that BBI produces ranges similar to the Monte Carlo estimates.

Figure \ref{fig:dtgoup} (right) shows results in \texttt{Go-Right-10}. Even the more expressive ``Sufficient'' model fails catastrophically, as do both Monte Carlo methods. The BBI agent still outperforms Q-learning.

\subsection{Feed-Forward Neural Networks}

A feed-forward neural network has multiple layers of {\em units}. A unit $u_{ij}$ in layer $i$ has a linear model $\hat{g}_{ij}$ and a monotonic {\em activation function} $c_{ij}$. Given input $x$, the output of $u_{ij}$ is $\hat{f}_{ij}(x) = c_{ij}(\hat{g}_{ij}(x))$. The input to the network is given as input to units in the first layer. Each subsequent layer is given the output of the previous layer as input. The output of the final layer is the output of the network.

We can iteratively perform output bound queries on the linear models in each row. Because $c_{ij}$ is monotonic, given an input bound $\overline{\underline{x}}$, we can let $\overline{f}_{ij}(\overline{\underline{x}}) = c_{ij}(\overline{g}_{ij}(\overline{\underline{x}}))$ and $\underline{f}_{ij}(\overline{\underline{x}}) = c_{ij}(\underline{g}_{ij}(\overline{\underline{x}}))$.

For outcome bound queries, we could maintain the extreme residuals as proposed for linear models. Alternatively, we can approximate the input-conditional maximum and minimum values with {\em quantile regression} via the pinball loss \cite{koenker1978regression}, which can be used to train the network to predict a high and low quantile. If the network outputs quantiles $\chi_{0.95}$ and $\chi_{0.05}$, then, given an input range $\overline{\underline{x}}$, we can approximate $\overline{f}(\overline{\underline{x}}) \approx \overline{\chi}_{0.95}(\overline{\underline{x}}))$ and $\underline{f}(\overline{\underline{x}}) \approx \underline{\chi}_{0.05}(\overline{\underline{x}}))$, the maximum possible value of the high quantile and the minimum possible value of the low quantile, respectively.

\subsubsection{Experiments}

\begin{figure}
  \centering
  \includegraphics[width=\linewidth]{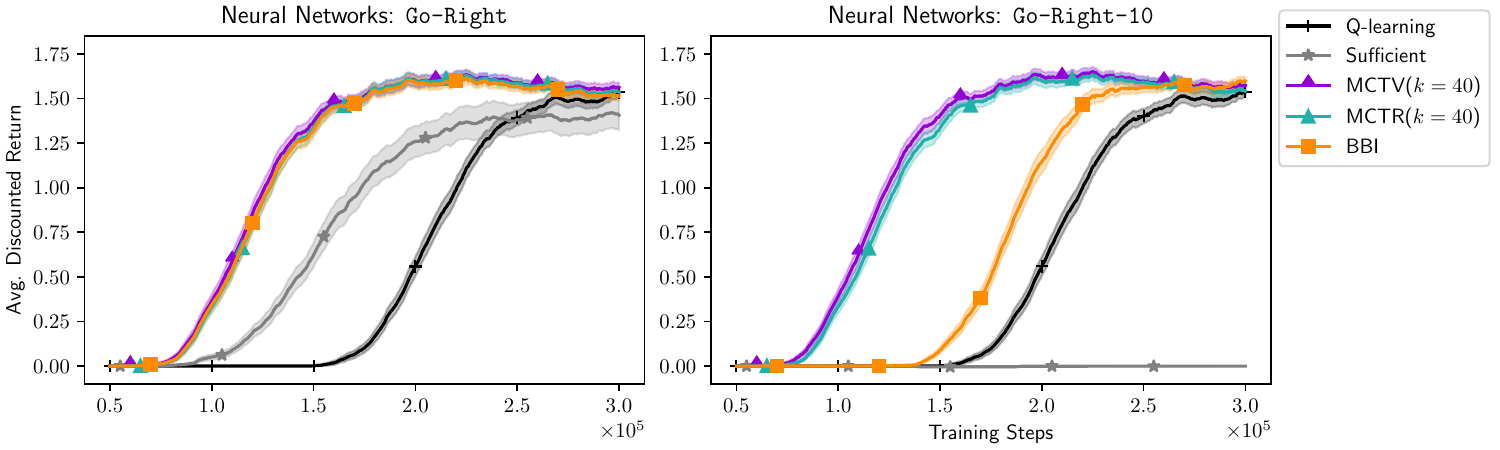}
  \caption{Selective planning with neural network models in \texttt{Go-Right} (left) and \texttt{Go-Right-10} (right).}
  \label{fig:nngoup}
\end{figure}

We trained 2-layer feed-forward neural networks using the ADAM optimizer \citep{kingma2015adam}. The first layer had 64 units with ReLU activation \citep{hahnloser2000digital}. The output layer used linear activation. For unselective planning, the model predicted the expected outcome. For BBI, the network additionally predicted $\chi_{0.05}$ and $\chi_{0.95}$. For sampling we trained an {\em implicit quantile network} (IQN) \citep{dabney2018implicit}, which uses quantile regression to learn arbitrary distributions. 

Figure \ref{fig:nngoup} shows results in \texttt{Go-Right} and \texttt{Go-Right-10}. In both cases the ``Sufficient'' model agent under-performs. BBI reliably outperforms Q-learning, but, in \texttt{Go-Right-10}, performs notably worse than in prior experiments. This suggests that the BBI uncertainties from the neural network may be more conservative than those from the hand-coded or regression tree models, resulting in less model usage. In Appendix \ref{app:nngor} we offer further discussion and empirical support of this hypothesis. BBI for neural networks may be prone to overly conservative planning since potentially loose bounds from one layer are given as input to the next layer, compounding any overestimates of uncertainty.

Monte Carlo methods perform better than predicted in \texttt{Go-Right-10}. In Appendix \ref{app:nngor}, however we argue that the results are consistent with our hypothesis that Monte Carlo inference is sensitive to the predicted distribution. In short, the particular errors in the neural network model create more variance in TD targets than in prior experiments, which is easily detected by Monte Carlo sampling.

Overall, from these results we observe that, given sufficient samples (relative to the model's predicted variance), Monte Carlo inference can provide more precise uncertainty estimates than BBI, which may suffer from overestimation of uncertainty.

\section{Experiments in \texttt{Acrobot}}
\label{sec:moreexp}

As a less contrived testbed, we consider the classic \texttt{Acrobot} control problem \citep{spong1989robot,dejong1994swinging}, which \citet{abbas2020selective} also explored. The agent controls the torque on one joint of a robot arm while the second is free swinging. The agent tries to swing the tip of the arm up to the level of the upper joint, receiving -1 reward until it does so. The discount factor $\gamma = 1$. We also introduce a variant called \texttt{Distractrobot} that has a distractor dimension sampled uniformly every step from $[-4\pi, 4\pi]$, the range of the upper joint's angular velocity.

\subsection{Experimental Setup}

In these experiments the agent will behave greedily, so planning will impact the agent's behavior (and therefore its own training data). This significantly complicates the interactions between the model and value learning. We truncate episodes after 500 episodes, if not terminated. 

The state-action value function $\hat{q}$ is linear over a tile coding feature set described by \citet{sutton1995generalization}, which we detail in Appendix \ref{app:tilecoding}. In a tile coding each tiling is a one-hot encoding, a fact that we exploit in output range queries on $\hat{q}$. We performed joint sweeps over both $\alpha$ and $\tau$ (details in Appendix \ref{app:sweep}) and ran 50 independent trials of each method with the selected metaparameter values.

To aid legibility of the results, we plot the difference between the average total reward of each method and that of the Q-learning agent. Positive values mean an agent is outperforming Q-learning. We provide more traditional learning curves in Appendix \ref{app:acrocurves}.

\subsection{Results}

\begin{figure}
  \centering
  \includegraphics[width=\linewidth]{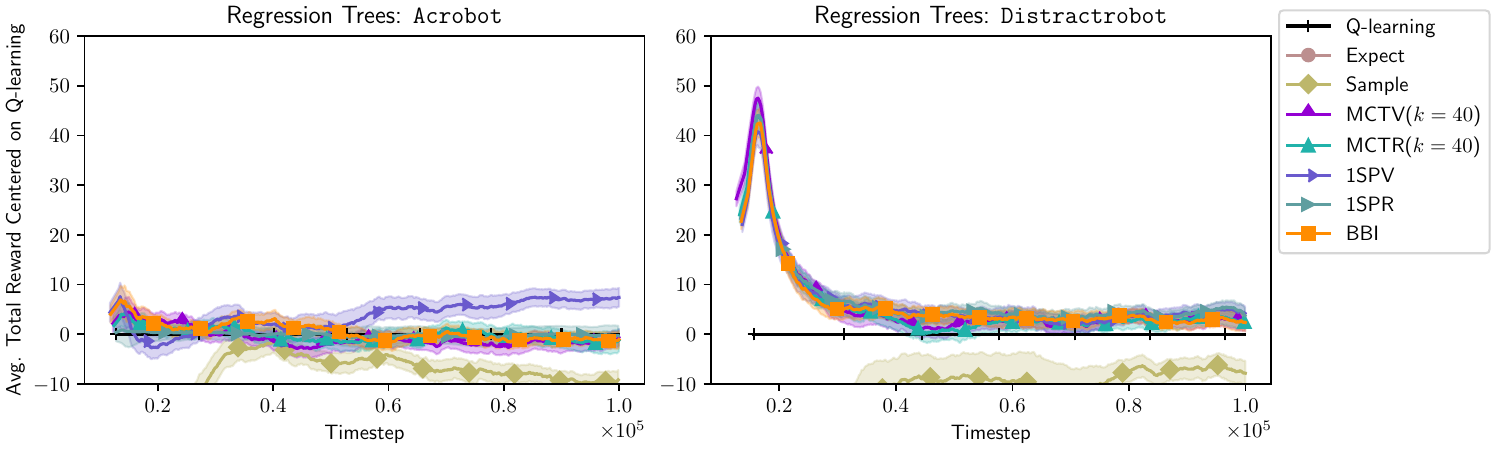}
  \caption{Planning with decision tree models in \texttt{Acrobot} (left) and \texttt{Distractrobot} (right). As above, curves are smoothed over 100 episodes and the shaded region represents standard error.}
  \label{fig:dtacro}
\end{figure}

Figure \ref{fig:dtacro} shows results in \texttt{Acrobot} and \texttt{Distractrobot} using regression trees learned using the FIRT algorithm. In order to limit the models' accuracy, we set a maximum of 100 leaves per tree. Both unselective planning agents (``Expect'' and ``Sample'') perform notably worse than Q-learning; their curves largely fall below the limits of these plots. All of the selective planning methods successfully mitigate the impact of the model's inaccuracies. In \texttt{Acrobot}, the 1SPV agent is the clear outlier; it outperforms all of the other agents, though it is unclear why. When additional state uncertainty is added in \texttt{Distractrobot}, it performs more modestly. In \texttt{Acrobot}, the other agents perform comparably to Q-learning. In \texttt{Distractrobot}, selective planning methods outperform Q-learning, especially early in training, with MCTV slightly outperforming the rest early on.

\begin{figure}
  \centering
  \includegraphics[width=\linewidth]{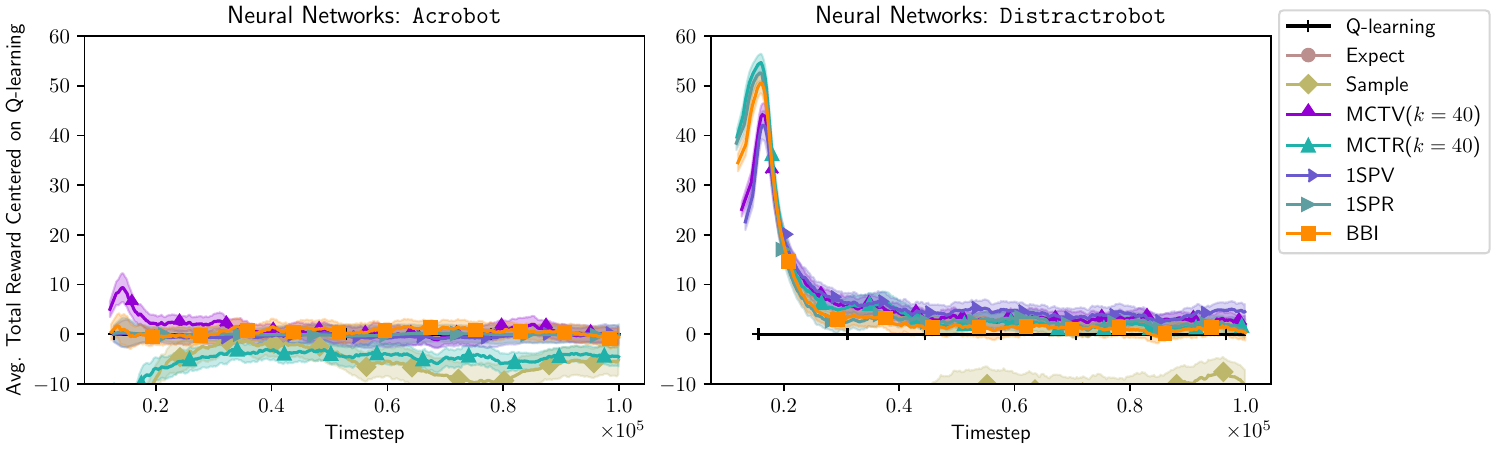}
  \caption{Planning with neural network models in \texttt{Acrobot} (left) and \texttt{Distractrobot} (right).}
  \label{fig:nnacro}
\end{figure}

Figure \ref{fig:nnacro} shows results using 2-layer feed-forward neural networks. The models have limited capacity with only 8 units in the first layer. As before, both unselective agents under-perform. The MCTV agent learns quickly in \texttt{Acrobot}, but the MCTR agent under-performs; perhaps the extreme values have low probability in the IQN model. The remaining selective planning methods perform comparably to Q-learning. In \texttt{Distractrobot}, the selective planning methods all learn more quickly than Q-learning, but the range-based methods perform notably better than the variance-based methods.

On the whole, the results are unsurprisingly not as clear as in \texttt{Go-Right}. In 3 out of the 4 experiments, inference-based and/or range-based methods learn quickest, which is consistent with our hypotheses. Also, once again BBI consistently avoids planning failure and, for the most part, enables planning benefits comparable to Monte Carlo inference.

\section{Summary of Conclusions}

Using hand-coded models, we observed that uncertainty over the model's one-step predictions (1SPV and 1SPR) can become overly conservative when state uncertainty does not lead to TD target uncertainty. We conclude that, for best selective planning results, the model should support inference of uncertainty over learning updates (e.g. TD targets).

We also observed, in both hand-coded and learned models, that Monte Carlo inference (MCTV and MCTR) can be effective, but it is sensitive to the model's predicted probability distribution, which may be influenced by the model's structural assumptions and even the behavior policy. With insufficient samples relative to that distribution, underestimated uncertainty may allow model error to interfere with learning. Furthermore, we argued that even exact target variance may not be a reliable inadequacy measure, as an arbitrarily small target variance can still indicate significant model inadequacy. We conclude that, for robust selective planning results, distribution insensitive uncertainty measures (such as the range) are preferable.

Finally, guided by these insights, we introduced bounding-box inference to obtain distribution-insensitive bounds over the TD targets. BBI's main drawback is overestimation of uncertainty (and thus lower model usage). As such, it can be outperformed by methods that provide more accurate inference (such as MCTV and MCTR with sufficient samples). That said, BBI was also consistently robust to irrelevant state error, which significantly impacted one-step uncertainty methods, and low-variance predicted distributions, which significantly impacted Monte Carlo methods.

Overall, we conclude that models that support distribution-insensitive uncertainty inference may be particularly desirable for selective planning, and that BBI is a promising step in that direction.

\section{Future Directions}

Selective MVE with bounding-box inference is an encouraging step toward MBRL agents that are less brittle in the face of model inaccuracy. Here we briefly describe some particularly important avenues for building upon this approach.

MVE is particularly amenable to selective planning but there are other important forms of planning. For instance, it is common to perform one-step value/policy updates along a model rollout. How should we perform updates on the values of simulated states, which may themselves be uncertain?

Bounding-box inference is appealing in its simplicity but the inferred bounds may be loose. Are there more complex representations of uncertainty that could support efficient, sound inference that accounts for some relationships between state variables and thus provides tighter bounds?

We rely on learned outcome bound estimates, but these may be based on insufficient data or poor generalization. We have focused on model inadequacy, but it will be important to integrate methods for detecting and mitigating epistemic uncertainty in the model and the uncertainty estimates.

TD target error is closely related to the quality of planning updates, but it is still a proxy measure. Some target error can be benign (for instance, causing a somewhat bigger update in the correct direction). How can we even more directly distinguish between helpful and harmful updates?

\subsubsection*{Acknowledgments}

This material is based upon work supported by the National Science Foundation under Grant No. IIS1939827 and by Harvey Mudd College. The experiments utilized GNU Parallel \citep{Tange2011a}. 

\bibliography{target_uncertainty_rlc2024}
\bibliographystyle{rlc}

\appendix
\section{Go-Right Details}
\label{app:goright}

The agent's underlying position in the hallway is represented as an integer in $\{0,1,\ldots, 10\}$. The agent observes its position with a uniformly randomly selected offset in $[-0.25, 0.25]$. The offset is sampled every time \texttt{Go-Right} is reset, and otherwise remains constant. The agent has two actions, $left$ and $right$, which decrease and increase its position, respectively. Taking the $left$ action in position 0 or the $right$ action in position 10 results in no change in position.

The status indicator light has 3 discrete underlying values representing it's intensity: $\{0, 5, 10\}$. The agent observes the light's intensity with an offset in $[-1.25, 1.25]$. The light's intensity at each step follows a 2nd-order Markov pattern with the following transition table:

\begin{table}[H]
  \centering
  \begin{tabular}{ c c | c}
    $t-1$ & $t$ & $t+1$ \\ \hline
    0 & 0 & 5 \\
    0 & 5 & 0 \\
    0 & 10 & 5 \\
    5 & 0 & 10 \\
    5 & 5 & 10 \\
    5 & 10 & 10 \\
    10 & 0 & 0 \\
    10 & 5 & 5 \\
    10 & 10 & 0 \\
  \end{tabular}
\end{table}

This pattern ensures that all intensities appear at equal frequencies and furthermore that each intensity is followed by each other intensity with equal frequency.

The prize indicator lights have 2 discrete underlying values representing their intensities: $\{0, 1\}$. The agent observes each light's intensity with independently sampled offsets in $[-0.25, 0.25]$. When the agent is not in position 10, all prize indicators have intensity 0. If the agent is in position 9 at time $t-1$ and at time $t$ in position 10 with the status indicator at intensity 10, all prize indicators transition to intensity 1 and remain at that value as long as the agent remains in position 10. When the agent enters position 10 with the status indicator at intensity 0 or 5, the prize indicators follow a pattern while the agent remains in position 10: all at intensity 0, only the left-most at intensity 1, only the next left-most at intensity 1, and so on until the right-most is at intensity 1 and then they all return to intensity 0 to repeat the pattern. This pattern ensures that the prize indicator lights are not fully identical in their behavior and must thus all be modeled. By default \texttt{Go-Right} has 2 prize indicators, but \texttt{Go-Right-10} has 10 prize indicators.

Taking the $left$ action results in 0 reward. When the agent is in position 10 with all prize indicators at intensity 1, taking the $right$ action results in 3 reward. In all other circumstances, taking the $right$ action incurs -1 reward. This ensures that going right is optimal but that the agent will initially learn to go left instead.

\section{Parameter Sweep Details}
\label{app:sweep}

In \texttt{Go-Right} and \texttt{Go-Right-10} we performed a grid search over $\alpha \in \{\num{1e-2}, \num{5e-2}, \num{1e-1}, \num{2e-1}\}$ and $\tau \in \{\num{1e-3}, \num{1e-2}, \num{1e-1}, \num{1}, \num{1e1}\}$ 

In \texttt{Acrobot} and \texttt{Distractrobot} we performed a grid search over $\alpha \in \{\num{5e-2}, \num{1e-1}, \num{2e-1}, \num{5e-1}\}$ and $\tau \in \{\num{1e-2}, \num{1e-1}, \num{1}, \num{1e1}, \num{1e2}\}$ 

For each configuration we ran 10 independent trials and calculated the {\em final performance} as the average discounted return of the last 100 episodes. For Q-learning, we selected the value of $\alpha$ that resulted in the highest final performance. For all other methods we compared the final performance of each configuration to Q-learning's (with the best $\alpha$ value). If no configuration obtained a higher final performance than Q-learning, we selected the configuration with the highest final performance. Otherwise, for each configuration with a final higher performance higher than Q-learning's, we calculated the difference between the score at each point and Q-learning's score at each point as a measure of the improvement from planning. We chose the configuration with the highest sum of these differences. The selected parameters are listed in Appendix \ref{app:selected}.

The results were generated by running 50 independent trials with each agent using the metaparameter values selected by this procedure.

\section{\texttt{Go-Right} with $\gamma = 0.85$}
\label{app:goleft}

It is possible to succeed in \texttt{Go-Right} by simply overestimating the value of going right rather than by correctly estimating the state-action values. We evaluated whether the selective planning methods, and BBI in particular, might be erroneously optimistic by applying them to \texttt{Go-Right} and \texttt{Go-Right-10} with the discount factor $\gamma = 0.85$. With the lower discount factor, the potential prize at the end of the hallway is not worth the cost to reach it. The optimal policy is to go left forever, collecting 0 reward. 

The following figures show planning performance with hand-coded, regression tree, and neural network models. In all cases, the selective planning methods learn to exclusively go left resulting in a discounted return of 0. In the case of neural networks, we do see that unselective planning with the sufficient model occasionally overestimates the value of going right and obtains negative discounted returns, but the selective planning methods, including BBI, avoid this misstep.

\begin{figure}[H]
  \centering
  \includegraphics[width=\linewidth]{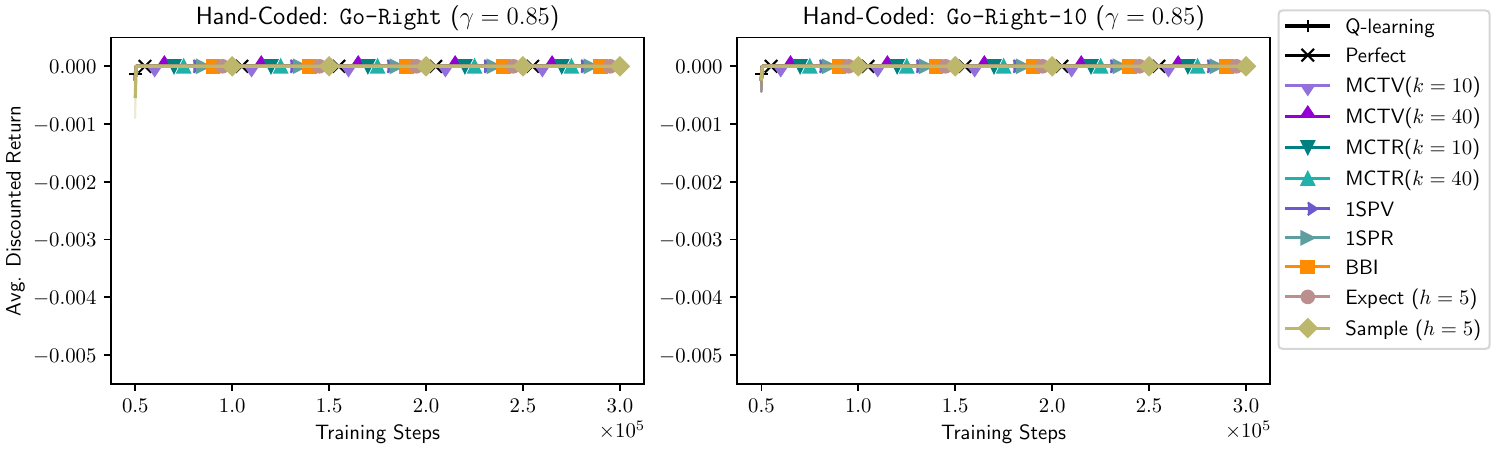}
  \caption{Hand-coded models in \texttt{Go-Right} and \texttt{Go-Right-10} with reduced discount factor.}
\end{figure}

\begin{figure}[H]
  \centering
  \includegraphics[width=\linewidth]{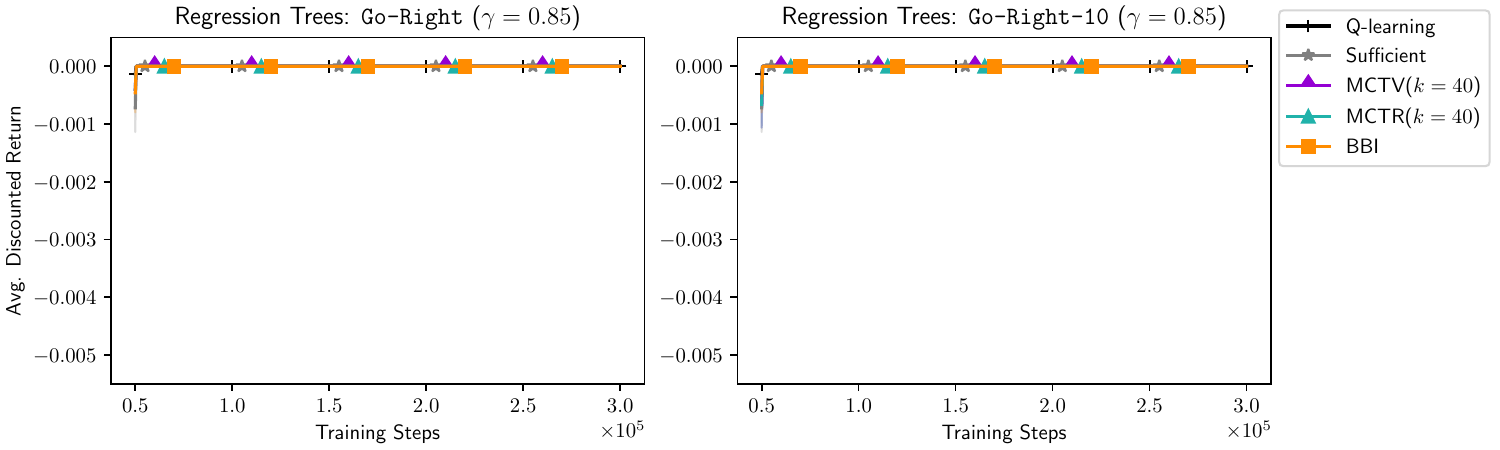}
  \caption{Regression tree models in \texttt{Go-Right} and \texttt{Go-Right-10} with reduced discount factor.}
\end{figure}

\begin{figure}[H]
  \centering
  \includegraphics[width=\linewidth]{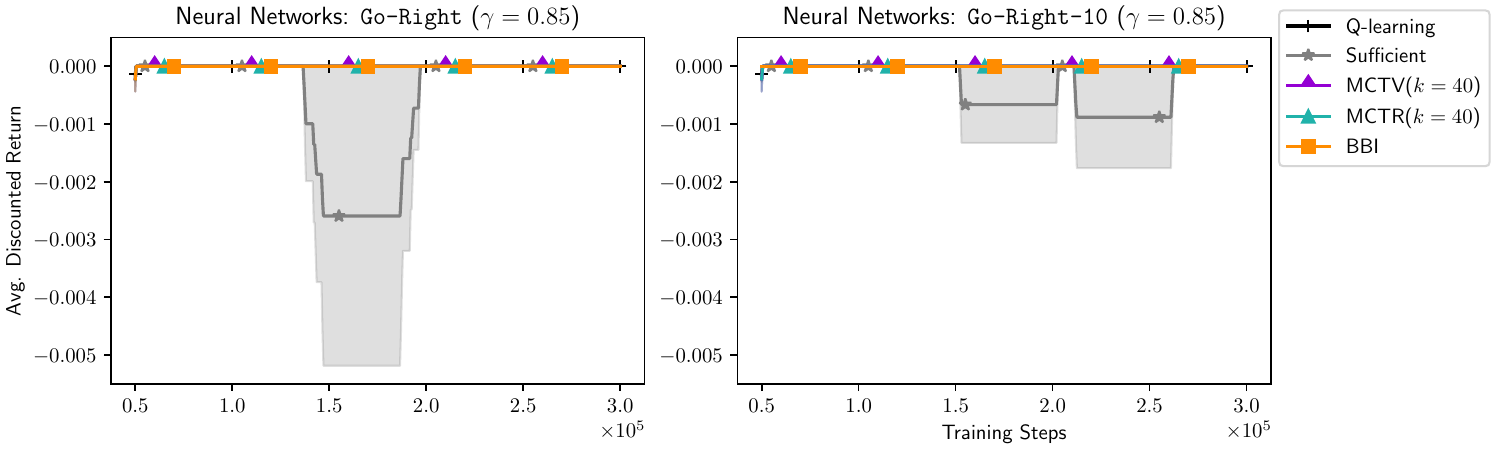}
  \caption{Neural network models in \texttt{Go-Right} and \texttt{Go-Right-10} with reduced discount factor.}
\end{figure}

\section{Model-Learning Details}
\label{app:metaparams}

We did not formally sweep metaparameters for the model-learning algorithms. Reasonably effective metaparameter values were selected informally during agent development.

The FIRT algorithm has two main metaparameters. First, there is a confidence level that must be reached in order to perform a split. We set this value to 0.95 (meaning a 0.95 chance that the best feature is, in fact, better than the second-best). Second, there is a threshold for the Hoeffding bound, below which best 2 features are considered tied and a split can be made. We set this value to 0.05. We incorporated every observed transition into the leaf statistics, but we only considered splitting the tree every 100 frames.

Every 4 steps we performed a training step on each neural network with a batch of 4 uniformly randomly selected transitions from a buffer containing all past transitions. We set the ADAM stepsize parameter to \num{1e-3}. For the BBI networks that predict the expected outcome as well as two quantiles, we weighed the three loss functions equally. 

\section{Additional Discussion of \texttt{Go-Right-10} with Neural Networks}
\label{app:nngor}

In Figure \ref{fig:nngoup} we observed that in \texttt{Go-Right-10}, MCTV and MCTR planning with neural network models did not fail as they did with hand-coded and regression tree models. Furthermore, we observed that the performance of BBI planning notably declined, unlike with hand-coded and regression tree models. In this section we aim to explain this observation in more detail.

We hypothesize that BBI with the neural network produces more conservative uncertainty estimates than with the hand-coded or regression tree models. To evaluate this, we consider the {\em uncertainty error}, the difference between the uncertainty estimate given by an inference procedure on a learned model $u^{learned}_i$ and the uncertainty estimate given using BBI with a hand-coded model $u^{hBBI}_i$: $\epsilon_{unc} = u^{learned}_i - u^{hBBI}_i$. For instance, if $\epsilon_{unc} > 0$, then the estimated uncertainty from the learned model is larger than the ideal BBI estimate. In each episode, we calculated the median uncertainty error over all planning steps and all rollout steps within a planning step. The below figure shows the median uncertainty error averaged over all trials for MCTV, MCTR, and BBI in both regression tree and neural network models.

\begin{figure}[H]
  \centering
  \includegraphics[width=\linewidth]{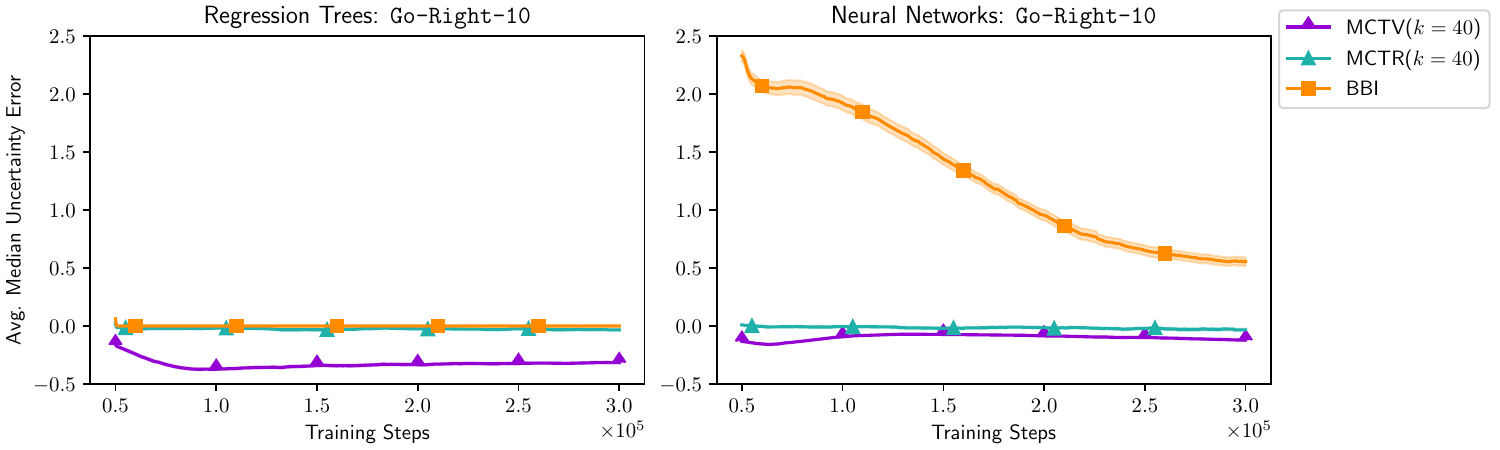}
  \caption{Median uncertainty error (averaged over all trials) in regression trees and neural networks.}
\end{figure}

We see that BBI in the neural network models frequently overestimates uncertainty, but not in the regression tree models. We suspect that this is because our BBI procedure for neural networks is additive (in that we assume that all inputs achieve their extreme values in the linear calculation in each unit) and iterative (in that the potentially loose bounds from each unit are given as input to the next layer of units). Both of these can compound the effects of loose bounds. The higher uncertainty estimates given by the neural network models may explain the reduced benefit from BBI planning observed in \texttt{Go-Right-10}.

We also observe that MCTV and MCTR planning in regression tree models frequently underestimate uncertainty (MCTV seemingly more so than MCTR). This intuitively coincides with the poor planning performance of these methods; this overconfidence may cause the planning process to use the model when it should not be trusted.

In the neural network models, the Monte Carlo methods seem to underestimate uncertainty to a lesser degree and perform well. Examining model rollouts provides a possible explanation for this. The neural network model generally fails to learn that the prize indicators may turn on when the agent enters the prize location (this is a relatively rare occurrence in the training set). This flaw makes the model highly misleading, which is borne out by the poor performance of the sufficient neural network model in this problem. However, as a side effect of this large error, samples from the IQN model produce high target variance, which is easily detected by the Monte Carlo method.

Specifically, when the agent enters the prize location, with roughly $1/3$ chance the status indicator has full intensity, but the prize indicators are off. This state is extremely rare in the training data, and thus has $q$-values close to initialization (i.e. close to 0). On the other hand, with $2/3$ chance, the status indicator does not have full intensity and the prize indicators are off; these are common states with relatively high $q$-values (since the agent can simply move left and then move right to try for the prize again). As such, there is high variance over TD targets that include the agent entering the prize location, which would otherwise be the main source of misleading updates

In contrast, the hand-coded and regression tree models both predict that the prize indicators will turn on with probability roughly $1/3$ when the agent enters the prize location. Thus, with high probability they predict that some, but not all, of the prize indicators will turn on. These are mostly impossible states that tend to all have $q$-values close to initialization, resulting in low variance in the TD targets, which is more difficult to detect via Monte Carlo sampling.

So, though these results were counter to our predictions based on the hand-coded model experiments, they are still consistent with our overall hypotheses. Predicted variance and sampling-based methods are sensitive to the model's predicted probability distribution; in this case the distribution happened to be amenable to these methods. BBI is insensitive to the model's predicted distribution but can generate loose bounds, erring on the side of conservative selective planning.

\section{Tile Coding Details}
\label{app:tilecoding}

For \texttt{Acrobot} we followed \citet{sutton1995generalization}. We had 12 tilings over all 4 dimensions, 3 tilings for each subset of 3 dimensions, 2 tilings for each subset of 2 dimensions, and 3 tilings for individual dimension. This gives a total of 60 tilings. The first and third dimensions were divided into 6 cells, the second and fourth into 7 cells.

For \texttt{Distractrobot} we extended the feature set to include the extra distractor dimension. We had 20 tilings over all 5 dimensions, 4 tilings over each subset of 4 dimensions, 2 tilings over each subset of 3 dimensions, 2 tilings over each subset of 2 dimensions, and 4 tilings over each individual dimension. This gives a total of 140 tilings. The first four dimensions were divided as above. The fifth dimension was divided into 7 cells.

\section{Acrobot Learning Curves}
\label{app:acrocurves}

Here we present more traditional learning curves for \texttt{Acrobot} and \texttt{Distractrobot} (i.e. not centered on Q-learning's performance).

\begin{figure}[H]
  \centering
  \includegraphics[width=\linewidth]{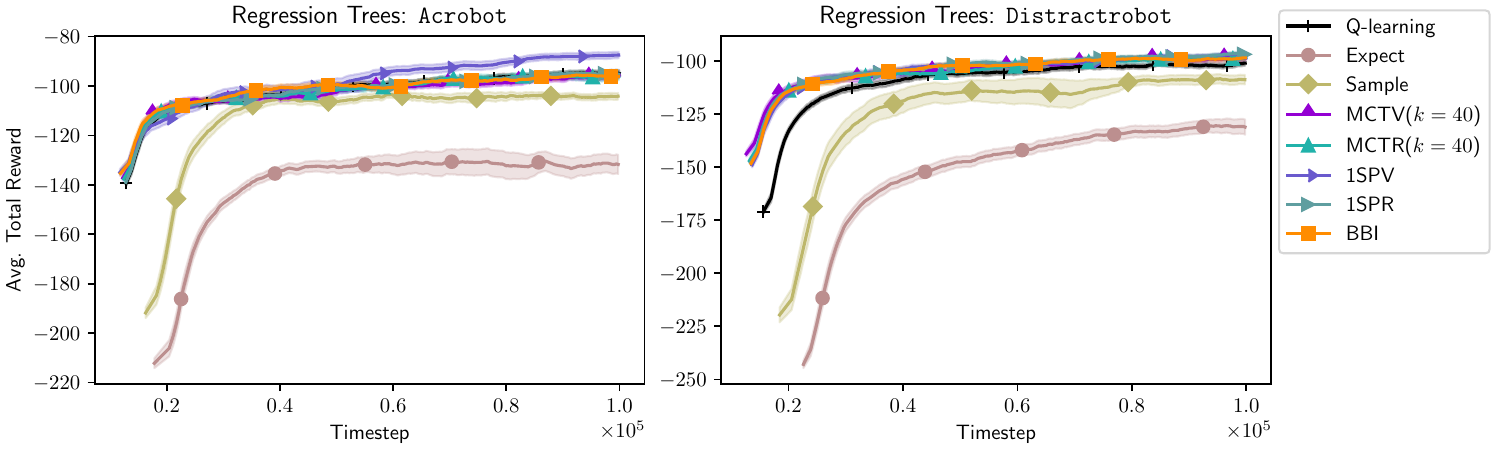}
  \caption{Planning with neural network models in \texttt{Acrobot} (left) and \texttt{Distractrobot} (right).}
\end{figure}

\begin{figure}[H]
  \centering
  \includegraphics[width=\linewidth]{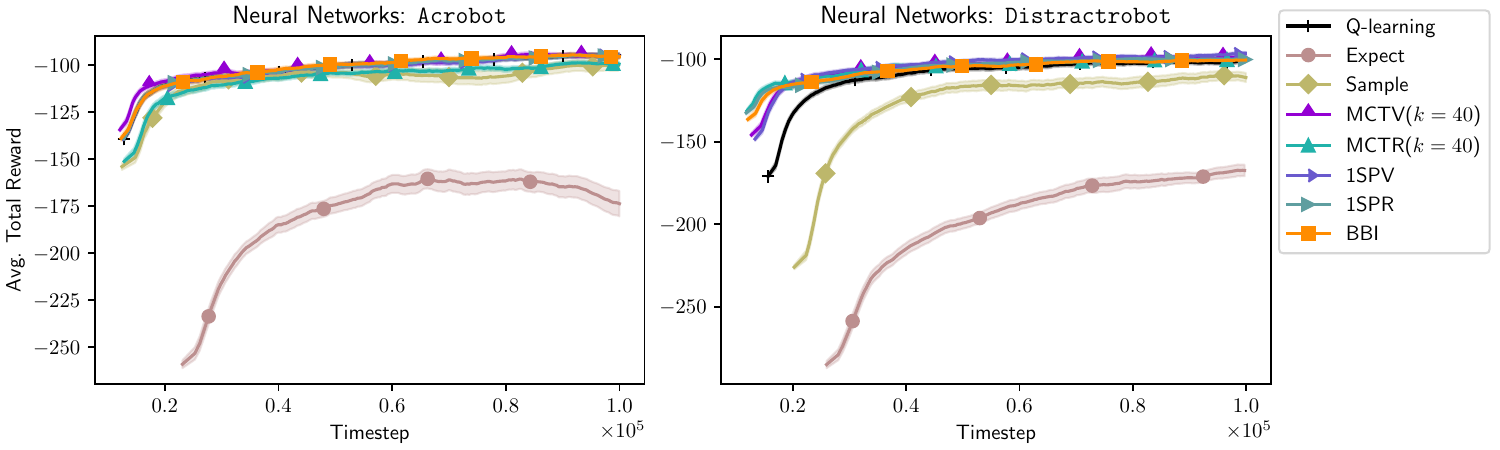}
  \caption{Planning with neural network models in \texttt{Acrobot} (left) and \texttt{Distractrobot} (right).}
\end{figure}

\section{Selected Metaparameters}
\label{app:selected}

Here we list the selected metaparameters for each agent in each problem. See Appendix \ref{app:metaparams} for details about how these values were selected.

\begin{multicols}{2}
  \begin{table}[H]
    \centering
    \caption{Hand-coded, \texttt{Go-Right} ($\gamma = 0.9$).}
    \begin{tabular}{ l | c | c}
      Agent & $\alpha$ & $\tau$ \\ \hline
      Q-learning & \num{5e-1} & n/a \\
      Perfect & \num{1e-1} & n/a \\
      Expect ($h = 2$) & \num{1e-1} & n/a \\
      Expect ($h = 5$) & \num{2e-1} & n/a \\
      Sample ($h = 2$) & \num{2e-1} & n/a \\
      Sample ($h = 5$) & \num{2e-1} & n/a \\
      1SPV & \num{5e-2} & \num{1e-1} \\
      1SPR & \num{5e-2} & \num{1} \\
      MCTV ($k = 10$) & \num{1e-1} & \num{1e-3} \\
      MCTV ($k = 40$) & \num{1e-1} & \num{1e-1} \\
      MCTR ($k = 10$) & \num{1e-1} & \num{1e-2} \\
      MCTR ($k = 40$) & \num{1e-1} & \num{1e-1} \\
      BBI & \num{1e-1} & \num{1}
    \end{tabular}
  \end{table}

  \begin{table}[H]
    \centering
    \caption{Regression tree, \texttt{Go-Right} ($\gamma = 0.9$).}
    \begin{tabular}{ l | c | c}
      Agent & $\alpha$ & $\tau$ \\ \hline
      Sufficient & \num{1e-1} & n/a \\
      MCTV ($k = 40$) & \num{1e-1} & \num{1e-2} \\
      MCTR ($k = 40$) & \num{1e-1} & \num{1e-1} \\
      BBI & \num{1e-1} & \num{1e-1}
    \end{tabular}
  \end{table}

  \begin{table}[H]
    \centering
    \caption{Neural net, \texttt{Go-Right} ($\gamma = 0.9$).}
    \begin{tabular}{ l | c | c}
      Agent & $\alpha$ & $\tau$ \\ \hline
      Sufficient & \num{1e-1} & n/a \\
      MCTV ($k = 40$) & \num{1e-1} & \num{1e-3} \\
      MCTR ($k = 40$) & \num{1e-1} & \num{1e-2} \\
      BBI & \num{1e-1} & \num{1e-1}
    \end{tabular}
  \end{table}
\end{multicols}

\begin{multicols}{2}
  \begin{table}[H]
    \centering
    \caption{Hand-coded, \texttt{Go-Right-10} ($\gamma = 0.9$).}
    \begin{tabular}{ l | c | c}
      Agent & $\alpha$ & $\tau$ \\ \hline
      Q-learning & \num{5e-2} & n/a \\
      Perfect & \num{1e-1} & n/a \\
      Expect ($h = 2$) & \num{2e-1} & n/a \\
      Expect ($h = 5$) & \num{2e-1} & n/a \\
      Sample ($h = 2$) & \num{2e-1} & n/a \\
      Sample ($h = 5$) & \num{2e-1} & n/a \\
      1SPV & \num{5e-2} & \num{1e-3} \\
      1SPR & \num{5e-2} & \num{1} \\
      MCTV ($k = 10$) & \num{2e-1} & \num{1e1} \\
      MCTV ($k = 40$) & \num{5e-1} & \num{1e-3} \\
      MCTR ($k = 10$) & \num{2e-1} & \num{1e1} \\
      MCTR ($k = 40$) & \num{5e-1} & \num{1e-1} \\
      BBI & \num{1e-1} & \num{1e-1}
    \end{tabular}
  \end{table}

  \begin{table}[H]
    \centering
    \caption{Regression tree, \texttt{Go-Right-10} ($\gamma = 0.9$).}
    \begin{tabular}{ l | c | c}
      Agent & $\alpha$ & $\tau$ \\ \hline
      Sufficient & \num{5e-1} & n/a \\
      MCTV ($k = 40$) & \num{2e-1} & \num{1e-1} \\
      MCTR ($k = 40$) & \num{2e-1} & \num{1} \\
      BBI & \num{1e-1} & \num{1e-1}
    \end{tabular}
  \end{table}

  \begin{table}[H]
    \centering
    \caption{Neural net, \texttt{Go-Right-10} ($\gamma = 0.9$).}
    \begin{tabular}{ l | c | c}
      Agent & $\alpha$ & $\tau$ \\ \hline
      Sufficient & \num{1e-1} & n/a \\
      MCTV ($k = 40$) & \num{1e-1} & \num{1e-3} \\
      MCTR ($k = 40$) & \num{1e-1} & \num{1e-2} \\
      BBI & \num{5e-2} & \num{1}
    \end{tabular}
  \end{table}
\end{multicols}

\begin{multicols}{2}
  \begin{table}[H]
    \centering
    \caption{Hand-coded, \texttt{Go-Right} ($\gamma = 0.85$).}
    \begin{tabular}{ l | c | c}
      Agent & $\alpha$ & $\tau$ \\ \hline
      Q-learning & \num{2e-1} & n/a \\
      Perfect & \num{2e-1} & n/a \\
      Expect ($h = 2$) & \num{2e-1} & n/a \\
      Expect ($h = 5$) & \num{2e-1} & n/a \\
      Sample ($h = 2$) & \num{2e-1} & n/a \\
      Sample ($h = 5$) & \num{2e-1} & n/a \\
      1SPV & \num{2e-1} & \num{1e1} \\
      1SPR & \num{2e-1} & \num{1e1} \\
      MCTV ($k = 10$) & \num{2e-1} & \num{1e1} \\
      MCTV ($k = 40$) & \num{2e-1} & \num{1e1} \\
      MCTR ($k = 10$) & \num{2e-1} & \num{1e1} \\
      MCTR ($k = 40$) & \num{2e-1} & \num{1e1} \\
      BBI & \num{2e-1} & \num{1e1}
    \end{tabular}
  \end{table}

  \begin{table}[H]
    \centering
    \caption{Regression tree, \texttt{Go-Right} ($\gamma = 0.85$).}
    \begin{tabular}{ l | c | c}
      Agent & $\alpha$ & $\tau$ \\ \hline
      Sufficient & \num{2e-1} & n/a \\
      MCTV ($k = 40$) & \num{2e-1} & \num{1e1} \\
      MCTR ($k = 40$) & \num{2e-1} & \num{1e1} \\
      BBI & \num{2e-1} & \num{1e1}
    \end{tabular}
  \end{table}

  \begin{table}[H]
    \centering
    \caption{Neural net, \texttt{Go-Right} ($\gamma = 0.85$).}
    \begin{tabular}{ l | c | c}
      Agent & $\alpha$ & $\tau$ \\ \hline
      Sufficient & \num{1e-1} & n/a \\
      MCTV ($k = 40$) & \num{5e-2} & \num{1e1} \\
      MCTR ($k = 40$) & \num{2e-1} & \num{1e-2} \\
      BBI & \num{2e-1} & \num{1}
    \end{tabular}
  \end{table}
\end{multicols}

\begin{multicols}{2}
  \begin{table}[H]
    \centering
    \caption{Hand-coded, \texttt{Go-Right-10} ($\gamma = 0.85$).}
    \begin{tabular}{ l | c | c}
      Agent & $\alpha$ & $\tau$ \\ \hline
      Q-learning & \num{2e-1} & n/a \\
      Perfect & \num{2e-1} & n/a \\
      Expect ($h = 2$) & \num{2e-1} & n/a \\
      Expect ($h = 5$) & \num{2e-1} & n/a \\
      Sample ($h = 2$) & \num{2e-1} & n/a \\
      Sample ($h = 5$) & \num{2e-1} & n/a \\
      1SPV & \num{2e-1} & \num{1e1} \\
      1SPR & \num{2e-1} & \num{1e1} \\
      MCTV ($k = 10$) & \num{2e-1} & \num{1e1} \\
      MCTV ($k = 40$) & \num{2e-1} & \num{1e1} \\
      MCTR ($k = 10$) & \num{2e-1} & \num{1e1} \\
      MCTR ($k = 40$) & \num{2e-1} & \num{1e1} \\
      BBI & \num{2e-1} & \num{1e1}
    \end{tabular}
  \end{table}

  \begin{table}[H]
    \centering
    \caption{Regression tree, \texttt{Go-Right-10} ($\gamma = 0.85$).}
    \begin{tabular}{ l | c | c}
      Agent & $\alpha$ & $\tau$ \\ \hline
      Sufficient & \num{2e-1} & n/a \\
      MCTV ($k = 40$) & \num{2e-1} & \num{1e1} \\
      MCTR ($k = 40$) & \num{2e-1} & \num{1e1} \\
      BBI & \num{2e-1} & \num{1e1}
    \end{tabular}
  \end{table}

  \begin{table}[H]
    \centering
    \caption{Neural net, \texttt{Go-Right-10} ($\gamma = 0.85$).}
    \begin{tabular}{ l | c | c}
      Agent & $\alpha$ & $\tau$ \\ \hline
      Sufficient & \num{1e-2} & n/a \\
      MCTV ($k = 40$) & \num{2e-1} & \num{1e1} \\
      MCTR ($k = 40$) & \num{2e-1} & \num{1e1} \\
      BBI & \num{5e-2} & \num{1}
    \end{tabular}
  \end{table}
\end{multicols}

\begin{multicols}{2}
  \begin{table}[H]
    \centering
    \caption{Regression tree, \texttt{Acrobot}.}
    \begin{tabular}{ l | c | c}
      Agent & $\alpha$ & $\tau$ \\ \hline
      Q-learning & \num{2e-1} & n/a \\
      Perfect & \num{5e-1} & n/a \\
      Expect & \num{5e-2} & n/a \\
      Sample & \num{5e-2} & n/a \\
      1SPV & \num{5e-1} & \num{1} \\
      1SPR & \num{2e-1} & \num{1e-2} \\
      MCTV ($k = 40$) & \num{2e-1} & \num{1e-1} \\
      MCTR ($k = 40$) & \num{1e-1} & \num{1e-2} \\
      BBI & \num{2e-1} & \num{1e1}
    \end{tabular}
  \end{table}

  \begin{table}[H]
    \centering
    \caption{Neural net, \texttt{Acrobot}.}
    \begin{tabular}{ l | c | c}
      Agent & $\alpha$ & $\tau$ \\ \hline
      Expect & \num{5e-2} & n/a \\
      Sample & \num{2e-1} & n/a \\
      1SPV & \num{2e-1} & \num{1e-2} \\
      1SPR & \num{2e-1} & \num{1e-2} \\
      MCTV ($k = 40$) & \num{2e-1} & \num{1e-1} \\
      MCTR ($k = 40$) & \num{1e-1} & \num{1} \\
      BBI & \num{2e-1} & \num{1e1}
    \end{tabular}
  \end{table}
\end{multicols}
\clearpage
\begin{multicols}{2}
  \begin{table}[H]
    \centering
    \caption{Regression tree, \texttt{Distractrobot}.}
    \begin{tabular}{ l | c | c}
      Agent & $\alpha$ & $\tau$ \\ \hline
      Q-learning & \num{1e-1} & n/a \\
      Perfect & \num{2e-1} & n/a \\
      Expect & \num{5e-2} & n/a \\
      Sample & \num{5e-2} & n/a \\
      1SPV & \num{2e-1} & \num{1e-1} \\
      1SPR & \num{2e-1} & \num{1e1} \\
      MCTV ($k = 40$) & \num{2e-1} & \num{1} \\
      MCTR ($k = 40$) & \num{2e-1} & \num{1e-1} \\
      BBI & \num{2e-1} & \num{1e-1}
    \end{tabular}
  \end{table}

  \begin{table}[H]
    \centering
    \caption{Neural net, \texttt{Distractrobot}.}
    \begin{tabular}{ l | c | c}
      Agent & $\alpha$ & $\tau$ \\ \hline
      Expect & \num{5e-2} & n/a \\
      Sample & \num{5e-2} & n/a \\
      1SPV & \num{2e-1} & \num{1e-2} \\
      1SPR & \num{5e-1} & \num{1} \\
      MCTV ($k = 40$) & \num{2e-1} & \num{1e-2} \\
      MCTR ($k = 40$) & \num{5e-2} & \num{1} \\
      BBI & \num{2e-1} & \num{1e1}
    \end{tabular}
  \end{table}  
\end{multicols}

\end{document}